\documentclass[11pt,a4paper]{article}
\usepackage[hyperref]{eacl2021}
\usepackage{times}
\usepackage{latexsym}

\usepackage{graphicx}
\usepackage{subcaption}
\usepackage{amsmath}
\usepackage[show]{notes}
\newcommand{\ignore}[1]{}
\newcommand{\com}[1]{}

\usepackage{microtype}

\aclfinalcopy 
\def\aclpaperid{774} 

\title{Open-Mindedness and Style Coordination in Argumentative Discussions}

\author{Aviv Ben Haim \\
  Department of Software and \\
    Information Systems Engineering \\
    Ben-Gurion University \\
  \texttt{avivbenh@post.bgu.ac.il} \And
  Oren Tsur \\
  Department of Software and \\
        Information Systems Engineering \\
        Ben-Gurion University \\
  \texttt{orentsur@bgu.ac.il} \\}

\date{}

\begin{document}
\maketitle
\begin{abstract}
Linguistic accommodation is the process in which speakers adjust their accent, diction, vocabulary, and other aspects of language according to the communication style of one another. Previous research has shown how linguistic accommodation correlates with gaps in the power and status of the speakers and the way it promotes approval and discussion efficiency. In this work, we provide a novel perspective on the phenomena, exploring its correlation with the open-mindedness of a speaker, rather than to her social status. We process thousands of unstructured argumentative discussions that took place in Reddit's Change My View (CMV) subreddit, demonstrating that open-mindedness relates to the assumed role of a speaker in different contexts. On the discussion level, we surprisingly find that discussions that reach agreement present lower levels of accommodation.  
\end{abstract}

\section{Introduction}

\subsection{Linguistic Accommodation}
Accommodation in personal communication refers to the unconscious process in which a speaker changes (accommodates) her communicative behavior with respect to the communication partner.
The change can be manifested across many dimensions, e.g., posture \cite{condon1967segmentation}, nodding \cite{hale1984models}, pauses \cite{jaffe_feldstein_1970}, and linguistic style \cite{niederhoffer2002linguistic}. From the perspective of sociolinguistic and social-psychology, accommodation is argued to increase cognitive efficiency \cite{giles1982cognition}, provide approval and validation \cite{giles2008engaging}, and project the speaker's positive image \cite{infante1997building}.

In this work we focus on the accommodation of the linguistic style -- the usage of stylistic markers and function words, such as auxiliary verbs and prepositions. \citet{tausczik2010psychological} estimated that style and function words make up for 55\% of the words we use. These markers shape the conversation regardless of its topical content.  Analysis of function-word usage is common in many NLP tasks, e.g., gender detection \cite{koppel2002automatically, mukherjee2010improving, bortolato2016intertextual}, forensic linguistics \cite{juola2008authorship, boukhaled2015using, kestemont2014function} and personality type detection \cite{argamon2005lexical, argamon2009automatically, litvinova2016profiling}, among others.

Accommodation is not necessarily a symmetric process. One party in a communication can accommodate while the other party can either accommodate as well (converge), sustain her behavior, or actively diverge. Linguistic style accommodation was first studied quantitatively by \citet{niederhoffer2002linguistic}, analyzing the linguistic style accommodation in a small scale dataset. Large scale datasets and advanced statistical and algorithmic methods were explored by \citet{danescu2011mark,guo2015bayesian,muir2017linguistic}, among others. 
Differences in the social status of the speakers were found to dominate accommodation \cite{danescu2012echoes}. 

In this paper we explore linguistic accommodation from a novel perspective. We argue that the speaker's \emph{open-mindedness} drives her linguistic coordination. While the speaker's open-mindedness can be ``socially forced'' by power relations -- it is also inherent to one's character and her actively assumed social role. 

\subsection{Open Mindedness} \emph{``To be open-minded is... to be critically
receptive to alternative possibilities, to be willing to think again despite having formulated a view, and to be concerned to defuse any factors that constrain one’s thinking in predetermined ways``} \cite{hare2003ideal}. Open-mindedness is closely related to the concepts of dialogism \cite{bakhtin1981dialogic} and the development of dialogic agency \cite{parker2006public}, vital in a liberal and multi-cultural society. 
Open-mindedness is found to help in conflict resolution \cite{tjosvold1999grievance,tjosvold1998dealing}, boost creativity \cite{mitchell2006knowledge,keskin2006market}, increase rationality and neutrality, and play a significant role in the theory of education \cite{hare1993open}.

From a practical perspective, the significant role open-mindedness plays in a range of situations, from the negotiation table to political debates and the classroom, requires an efficient way to detect open-mindedness or the lack-of, allowing intervention by a moderator or a teacher \cite{zakharov2020discourse}. We show the level accommodation is simple to compute and can be used as a proxy for open-mindedness.

\subsection{Reddit's CMV}
\label{subsec:cmv}
The Change My View (CMV) is a subreddit (forum) on the forum-based Reddit platform. The forum is self described as \emph{``A place to post an opinion you accept may be flawed, in an effort to understand other perspectives on the issue. Enter with a mindset for conversation, not debate.''}\footnote{https://www.reddit.com/r/changemyview/wiki/index (accessed Oct. 5 2020)}. Each discussion thread in CMV evolves around the topic presented in the submission by the Original Poster (OP). A discussion, therefore, takes the structure of a tree with the \emph{submission} at the root and the various comments (replies) stemming from it. One unique feature of the CMV subreddit is the \emph{Delta} {$\Delta$} -- a way to acknowledge a convincing argument. A Delta can be awarded by any user, OP or not, to a comment made by any other user, and should be explicitly justified. A user can award a Delta only to users holding an opposing view to her own.  A delta can signify ``good point'' and does not have to reflect a complete reversal of opinion. CMV is heavily moderated to maintain a high level of discussion and to ensure that Deltas are awarded with proper reasoning. Table \ref{tab:submission_example} provides a partial and truncated example of a submission from CMV.

\begin{table*}[t]
\centering
\footnotesize
\renewcommand{\arraystretch}{1.5} 
\begin{tabular}{ c|p{11.5cm}|c } 
 {\bf Speaker} & {\bf Text} & {\bf Gave delta}\\
 \hline
 \hline
 $A$ & \textbf{CMV: Money needs to get out of politics} \newline
 I am genuinely curious about the counter arguments as I haven't heard any and I want to know if I'm wrong.
 This began with this case: Buckley v. Valeo, 424 U.S. 1 (1976), was a landmark decision of the US Supreme Court on campaign finance... & \\ \hline
  $B$ & $>$ ...There's a lot of attention given to the massive amount of money that come from wealthy individuals or special interest groups but ordinary people can make campaign contributions to the candidates they support, too. \newline 
  Bernie Sanders only accepted donations from small-dollar donors and outspent Joe Biden in primary at a whopping \$200 million... &  \\ \hline
 $A$ & $>>$ You gave a good counter argument. I meant more big money, but I see it's hard to separate the two. $\Delta$  & Yes \\ \hline
 $C$ &  $>$ There have actually been a lot of studies suggesting that political donations don't swing votes... & \\ \hline
\end{tabular}
\caption{Partial and truncated example of a submission from CMV}
\label{tab:submission_example}
\end{table*}

\paragraph{Open-mindedness in CMV} We use CMV discussions to study the relation between linguistic accommodation and open-mindedness. While open-mindedness is an informal prerequisite of all CMV participants, we use two explicit indications of open-mindedness: (i) Being an OP -- Stating an opinion and literally declaring the willingness to change a presupposition, and (ii) The use of $\Delta$, explicitly acknowledging a convincing argument made by an opposing party. \\
It could be argued that the OPs are not necessarily open-minded and only respond to people that agree with them. Fortunately, CMV rules strive to create a constructive conversation and CMV is a heavily moderated subreddit. The rule states an OP is obliged to reply to opposing views and show open-mindedness.

\paragraph{Dataset} Our corpus consists of 9,374 English discussions initiated by 4873 OPs between January 2018 and June 2020. It contains 1,301,545 posts (utterances) by 86,941 unique users. 8,659 Deltas were awarded, 877 of them by users that are not the OP.
The mean number of speakers in each submission is 42, max and min are 898 and 3 respectively, and the median is 26.
The mean number of comments in each submission is 138, max and min being 2559 and 26 respectively, and the median is 88.
The mean numbers of word in CMV comment is 83 and the median is 50.

\subsection{Research Questions}
\label{subsec:rq_intro}
Based on the indications described above, we explore the relations between open-mindedness and accommodation through the following research questions:

\paragraph{RQ1: Do OPs present higher levels of linguistic accommodation?} By definition, OPs are open to change their view, hence open-minded. We expect to observe higher levels of linguistic coordination presented by the OP of a discussion than by other participants.

\paragraph{RQ2: Do Delta givers present higher levels of linguistic accommodation?} Delta givers explicitly express open-mindedness by awarding a $\Delta$ upon recognizing convincing argumentation. We expect to observe higher levels of linguistic coordination presented by the Delta givers than by other participants.
In the scope of this research question, we will focus on delta givers that are not the OPs of the discussion in which they awarded the Delta. By doing so we examine delta givers that unlike OPs, have no status -- they are not the initiators of the discussion, are not entitled to give the `final verdict', and can choose to opt-in or out. Since previous work explains accommodation in terms of power differences, delta givers serve as a control group, lacking any formal power or privilege.

\paragraph{RQ3: Are Deltas correlated with accommodation on the \emph{discussion level}, rather than on the user level?} We hypothesize that Delta awards may be correlated with the general accommodation observed in a discussion, rather with the open-mindedness of a specific user -- the Delta giver. In order to test this hypothesis we compare the accommodation of all users in branches (of the discussion tree) in which Deltas were awarded, to the accommodation in branches in which no Delta was awarded.

The formal definitions of user, group and aggregate coordination are presented in Section \ref{subsec:formal_coordination}, and the hypotheses promoted by research questions above are formally defined in Section \ref{subsec:formal_rq} and tested in Section \ref{sec:results}. Further analysis and refined hypotheses are proposed and tested in Section \ref{subsec:extra_rq}.

\section{Related Work}
\label{sec:related}
Linguistic coordination, sometimes referred to as `accommodation' or `matching', is mainly studied from two perspectives -- social (-psychology) and computational. Linguistic style matching in social interactions and its correlation to various psychometric properties of language was first studied by \citet{niederhoffer2002linguistic}. This approach was later generalized to discussions in small groups by \cite{gonzales2010language}. These metrics were applied to social media data by \citet{danescu2011mark}.

\citet{mukherjee2012analysis} analyzed a large corpus of online debate, reporting that debaters using agreement expressions (e.g., `I agree with that point', `rightly said') as a debate strategy show higher linguistic accommodation than debaters that explicitly express disagreement (e.g., `I don't buy it', `I disagree', `nonsense'). However, the use of agreement/disagreement expressions in a formal debate setting is a strategic choice and does not imply actual agreement or open-mindedness. Moreover, the stated objective in a formal debate is winning the debate, rather than self-reflection through convincing argumentation.

Power differences were found to drive linguistic coordination toward the speaker with the higher status \cite{danescu2012echoes}, and a particular configurations of personal traits (e.g., agreeableness, Machiavellianism, low self-consciousness) combined with low-power position were shown to increase the likelihood of coordination \cite{muir2016characterizing,muir2017linguistic}.
\citet{guo2015bayesian} proposed Bayesian inference approach to capture influence manifested through linguistic accommodation. 

This work is the first to explore coordination through the lens of open-mindedness, rather than through self-awareness, power, influence or strategic manipulation.

\section{Measuring Linguistics Accommodation}
\label{sec:formal}

\subsection{Linguistic Style Coordination}
\label{subsec:formal_coordination}
We adopt the common coordination metric used in previous studies by  \citet{danescu2011mark,danescu2012echoes,mukherjee2012analysis} and others. The metric quantifies the accommodation of an individual along a linguistic category by looking at the shifting usage of terms of that category, compared to other speakers. The linguistic categories include Articles, Adverbs, Quantifiers, Conjunctions, Indefinite Pronouns, Personal Pronouns, Prepositions and Auxiliary Verbs as well as other psycho-linguistic categories from the widely used LIWC dictionary \cite{liwc}.

\paragraph{Speaker's accommodation} Given users $a$ and $b$, and a linguistic marker $m$, we want to measure how the usage of $m$ by $a$ triggers (or suppresses) the occurrence of $m$ in the direct responses of $b$. 

Given a set $S_{a,b}: (a: u_{1}, b: u_{2})$ of exchanges where $u_{1}$ denotes an utterance (comment) by $a$ and $u_{2}$ is a direct reply by $b$ we define coordination of $b$ towards $a$ on marker $m$ as:
\begin{equation}\label{eq:coordination}
C^m(b,a) = P(\xi^m_{u_2 \rightarrow u_1} | \xi^m_{u_1} ) - P(\xi^m_{u_2 \rightarrow u_1})
\end{equation}
Where $P(\xi^m_{u_1})$ is the probability of speaker $a$ using marker $m$;  $P(\xi^m_{u_2 \rightarrow u_1})$ is the probability of speaker $b$ using $m$ in replying to $u_1$ (made by $a$), and $P(\xi^m_{u_2 \rightarrow u_1} | \xi^m_{u_1} )$ is the probability of speaker $b$ using $m$ in replying to an utterance $u_1$ in which $a$ also used $m$. Equation \ref{eq:coordination} bears three important properties: 
\begin{enumerate}
    \item It is asymmetric, that is $C^{m}(b,a) \neq C^{m}(a,b)$, allowing us to realize which user is accommodating toward the other.
    \item $C^{m}(b,a)=0$ if any of the speakers uses $m$ in every utterance, since no turn-taking influence could be modeled in this scenario. 
    \item $C^{m}(b,a) \xrightarrow[|S^{\neg m}_{a,b}| \rightarrow \infty]{} 1$ if $b$ uses $m$ in a response to $a$ \emph{if and only if} $a$ used $m$, where $S^{\neg m}_{a,b}$ is the set of exchanges between $a$ and $b$ the do not contain $m$.
\end{enumerate}

\paragraph{Group Coordination} The definition of coordination is extended to address the coordination of a group of speakers. This enables us to measure (i) The coordination of a speaker $b$ toward a group of speakers $A$, and (ii) The coordination of a group $B$ toward a group of speakers $A$.

We accomplish (i) by simply looking at the set of consecutive utterances: $S_{A,b}: (A: u_{1}, b: u_{2})$  that includes the responses of speaker $b$ to every speaker $a \in A$. The coordination of $b$ toward the group $A$ is therefore defined as:

\begin{equation}\label{eq:coordination_single_to_group}
C^m(b,A) = P(\xi^m_{u_2 \rightarrow u_1} | \xi^m_{u_1} ) - P(\xi^m_{u_2 \rightarrow u_1})
\end{equation}

To accomplish (ii), the coordination of speakers in group $B$ to targets in group $A$, we calculate the average coordination of speakers in $B$ toward targets speakers $A$:

\begin{equation}\label{eq:coordination_group_to_group}
C^m(B,A) = \langle C^m(b,A) \rangle _{b \in B}
\end{equation}

\paragraph{Aggregated coordination} The coordination measures defined above apply to a specific marker $m$. Previous studies report that accommodation is not necessarily correlated across markers as user $b$ can accommodate toward $a$ on $m$ and diverge on $m'$ \cite{giles1982cognition,bilous1988dominance}. Therefore, having an aggregated coordination metric may provide a fuller picture regarding the overall accommodation across markers $M$ and groups.
Three types of \emph{aggregated} coordination measure are considered:

\begin{enumerate}
    \item \emph{Aggregate 1}: We first compute $C(b,A)$ as the macro average of $C^{m}(b,A)$ for all $m \in M$, then use it to compute $C(B,A)$ as described in Equation \ref{eq:coordination_group_to_group}. However, we wish to note that $C^{m}(b,A)$ can only be computed if there is a sufficient number of exchanges between $b$ and $A$ with the marker $m$ (we enforce $\delta > 3$), and therefore, \emph{Aggregate 1} is calculated \emph{only} using those $b \in B$ where $C^{m}(b,A)$ is defined for every $m$. This aggregation provides the most accurate generalized coordination, but relies only on a subset of the speakers. In order to compensate for cases in which $C^{m}(b,A)$ is undefined we propose the two aggregated measures below. 
    
    \item \emph{Aggregate 2}: Assuming group homogeneity, we approximate $C^{m}(b,A)$ by averaging $C^{m}(B, A)$ over all $b \in B$ for whom $C^{m}(b,A)$ is defined.
    
    \item \emph{Aggregate 3}: Assuming speakers tend to present the same behaviour across markers, the undefined $C^{m}(b,A)$ can be approximated by averaging  $C^{m'}(b,A)$ over all $m' \in M$ for which  $C^{m'}(b,A)$ could be computed directly. While this assumption does not always hold, this definition of aggregated coordination has proved useful in \citet{danescu2012echoes}.
\end{enumerate}

We used the ConvoKit toolkit for conversation analysis \cite{chang2020convokit} in preprocessing and measuring coordination.  

\begin{table*}[t]
\centering
\footnotesize
\renewcommand{\arraystretch}{1.5} 
\begin{tabular}{ c|p{11.5cm}|c } 
 {\bf Notation} & {\bf Definition} & {\bf Used in}\\
 \hline
 \hline
 $U$ & The set of all users (speakers) in the data & all \\ 
  $G^{OPs}$ & The set of all OPs  & $H_{1.1}$, $H_{1.2}$, $H_{4.3}$ \\ 
 $G^{\overline{OPs}}$ & The set of all users in a submission that are not the OP (AKA regulars)  & $H_{1.1}$, $H_{4.3}$ \\ 
 $G^{\overline{r}}$ & Regulars users in a submission that are OPs in other submissions & $H_{1.2}$ \\
    $G^{\Delta_{reg}}$ & The set of Delta givers $\notin G^{OPs}$. That is: Delta givers that are not the OP of the submission  & $H_2$, $H_{4.1}$ \\
    $G^{\Delta_{OP}}$ & The set of Delta givers $\in G^{OPs}$. That is: OPs that gave Delta in their own submission & $H_{4.2}$ \\
    $G^{\overline{\Delta}}$   &   The set of users that did not award a $\Delta$ in the submission  & $H_2$, $H_4$, $H_{4.1}$ \\
    $C^\Delta(\circ,\circ)$ & Coordination is measured only for discussion branches in which a $\Delta$ was awarded & $H_3$ \\
    $C^{\overline{\Delta}}(\circ,\circ)$ & Coordination is measured only for discussion branches in which a $\Delta$ was \emph{not} awarded & $H_3$ \\
    $G^\Delta$     & The set of Delta givers in a submission regardless of role & $H_4$ \\
    $G^{\overline{OP_\Delta}}$ & The set of OPs that did not give delta in their own submission  & $H_{4.2}$ \\
\end{tabular}
\caption{Notations and definitions for groups of speakers assuming different roles in the CMV data.}
\label{tab:notations}
\end{table*}

\subsection{Formalizing the Research Questions} 
\label{subsec:formal_rq}
\paragraph{Notation} Before using the definitions above to formalize the research questions presented in Section \ref{subsec:rq_intro}, we introduce a few more notations, presented in Table \ref{tab:notations} for convenience. An important aspect to note is that for each Reddit user $a$, we produce dummy users for each submission she takes part in. Formally: a user $a$ participating in  $n$ discussions will have $n$ dummy users $\{a_i\}_{i=1...n}$, one for each discussion. Different dummy representations of a user may be part of different groups, for example, in a discussion she initiated: $a_i \in G^{OPs}$ and in another discussion she participates (but didn't initiated) $a_j \in  G^{\overline{OPs}}$. These dummy representations allows us to explore the coordination of a user in different contexts. \\
We provide the following example of a conversation to familiarize with the notations: Given a submission opened by Bobby, comments by Jess, Arnold and a delta given by both Bobby and Ava. We can attribute our speakers to the groups accordingly: Bobby is in the group of OPs, the group of delta givers and the group of OPs that gave delta - $Bobby_1 \in G^{OPs}, G^\Delta, G^{\Delta_{OP}}$. Jess and Arnold are in the group of users that did not give delta - $\{Jess_1, Arnold_1\} \subseteq G^{\overline{\Delta}}$. Ava is in the group of delta givers and the group of delta givers that are not OPs - $Ava_1 \in G^\Delta, G^{\Delta_{reg}}$.

    \paragraph{RQ1: Do OPs present higher levels of linguistic accommodation?} In order  to answer positively we need to reject the null hypothesis:    \begin{equation}\label{eq:coordination_OP_vs_all}
    \tag{$H_{1.1}$}
    C(G^{OPs}, U) = C(G^{\overline{OPs}}, U)
    \end{equation}
    and verify that $C(G^{OPs}, U) > C(G^{\overline{OPs}}, U)$, showing that coordination of OPs (in discussions they initiated) towards all other users is  higher than the coordination of regular users toward all other users. 
    
    We further explore the behaviour of OPs putting forward the following null hypothesis:
    \begin{equation}
     \label{eq:coordination_OP_role}
     \tag{$H_{1.2}$}
    C(G^{OPs}, U) = C(G^{\overline{r}}, U)
    \end{equation}
    Namely, we test whether OPs present different levels of coordination in discussions they initiated and in discussions they participate without assuming the role of the OP.

    \paragraph{RQ2: Do Delta givers present higher levels of linguistic accommodation?} A positive answer is validated by rejecting the following null hypothesis:
    
    \begin{equation}\label{eq:coordination_delta_givers}
    \tag{$H_2$}
    C(G^{\Delta_{reg}}, U) = C(G^{\overline{\Delta}}, U)
    \end{equation}
    and verifying that $C(G^{\Delta_{reg}}, U) > C(G^{\overline{\Delta}}, U)$.
    
    It is important to note that $G^{\Delta_{reg}}$ only includes $\Delta$-givers that are not OPs (see definitions in Table \ref{tab:notations}). This distinction is important due to the fact that a user in an OP role declares her flexibility by inviting other users to change her mind, while other $\Delta$-givers do not assume this flexibility.

    \paragraph{RQ3: Are Deltas correlated with accommodation on the discussion level, rather than on the user level?} That is we check whether Deltas are awarded more generously in discussions with higher levels of accommodation, not necessarily by or toward the $\Delta$-giver. We put forward the following null hypothesis:
    \begin{equation}\label{eq:coordination_delta_branch}
    \tag{$H_3$}
    C^\Delta(U, U) = C^{\overline{\Delta}}(U, U)
    \end{equation}
   
    Validating $H_3$ shows that the general accommodation level is not correlated with $\Delta$ awards.

\section{Results and Analysis} 
\label{sec:results}
All of the hypothesis presented above are tested using the two-sample $t$-test. We report significance values for the following $p$-values: 0.05, 0.01 and 0.001. Conversations are extracted from the discussion data described in Section \ref{subsec:cmv}, containing 1.3M utterances by 87K unique users participating in 9.3K discussions.

\subsection{Open-mindedness}

\begin{figure*}[h!]
  \begin{minipage}[t]{0.45\textwidth}
  \vspace{0pt}
  \includegraphics[width=\columnwidth]{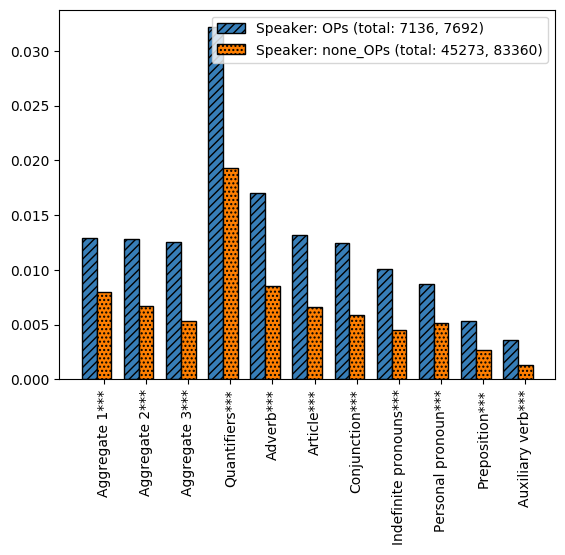}
  \subcaption{OP vs non-OP coordination to others. Testing \eqref{eq:coordination_OP_vs_all}}
  \label{fig:speaker_op}
  \end{minipage}\qquad
  \begin{minipage}[t]{0.45\textwidth}
  \vspace{0pt}
  \includegraphics[width=\columnwidth]{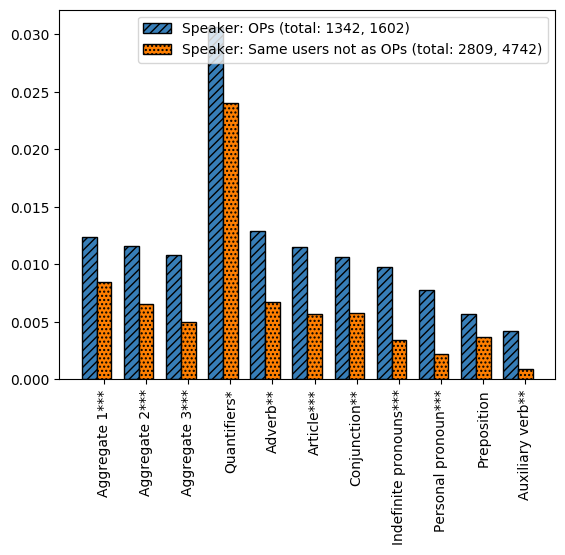}
  \subcaption{OP vs OP as regular participants coordination to others. Testing \eqref{eq:coordination_OP_role}}
    \label{fig:op_role_effect}
  \end{minipage}
  \caption{Parentheses in figure legends denote (number of speakers used in type 1 aggregation, number of speakers used in aggregations of type 2 and 3). Asterisks denote significance using independent t-test: */**/*** for $p < 0.05/ 0.01 / 0.001$, respectively.}
  \label{fig:rq1}
\end{figure*}

\paragraph{OPs present higher coordination (RQ1)} Figure \ref{fig:rq1} presents the coordination values for a number of marker types and for the three aggregate measures. We reject the null hypothesis $H_{1.1}$ with $p<0.001$ for all markers and measures and confirm that OPs present significantly higher levels of coordination, compared to regular users (Figure \ref{fig:speaker_op}). Similarly, we reject the null hypothesis $H_{1.2}$, though significance values vary between markers (Figure \ref{fig:op_role_effect}). 

The results above show that the open-mindedness of an OP is not an inherent personal capacity but depends on the role assumed in a specific context. We further validated this by testing the coordination of OPs in discussions they did not initiate ($G^{\overline{r}}$) to the coordination level of users that never initiated a discussion as OPs. Indeed, we find no significant difference between their coordination levels.

\begin{figure}[h!]
  \includegraphics[width=\columnwidth]{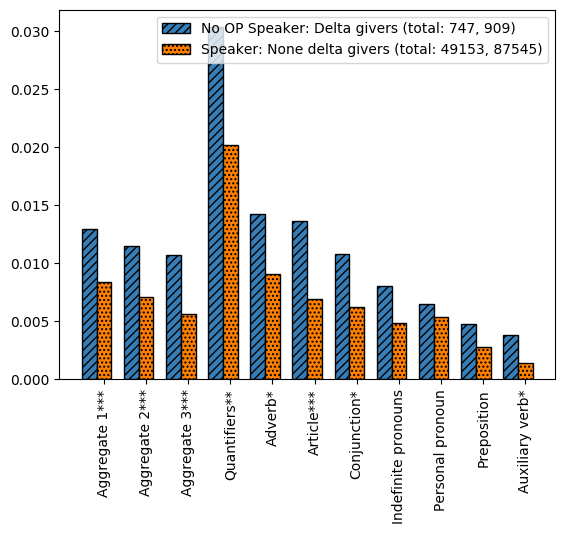}
  \caption{Non-OP Delta givers vs Users that did not give Delta. Test \eqref{eq:coordination_delta_givers}}
\label{fig:delta_givers}
\end{figure}

\paragraph{Delta givers present higher coordination (RQ2)} We reject the null hypothesis \eqref{eq:coordination_delta_givers} for most markers and all aggregate types (Figure \ref{fig:delta_givers}). This result confirms that non-OP Delta givers present higher levels of coordination compared to users that did not give Delta.

\begin{figure}[h!]
  \includegraphics[width=\columnwidth]{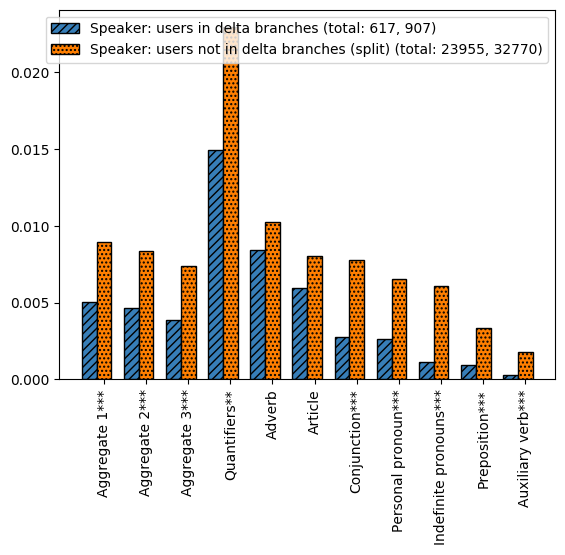}
  \caption{Delta branches vs Non-Delta branches. Test \eqref{eq:coordination_delta_branch}}
  \label{fig:delta_branch}
\end{figure}

\paragraph{Coordination varies on branch level (RQ3)} We reject the null hypothesis \eqref{eq:coordination_delta_branch} for most markers and all aggregate measures, asserting that coordination levels in branches in which a $\Delta$ was awarded are significantly different from those in branches without any $\Delta$. Surprisingly, we find that the levels of accommodation are higher in branches where no Delta was awarded (see figure \ref{fig:delta_branch}). In the next sub-section we further explore this surprising result.

\subsection{Accommodation Effect on Delta Giving}
\label{subsec:extra_rq}
In the previous section we have established that on the user-level, $\Delta$-givers present higher coordination than other users, however, on the discussion level, $\Delta$ is more likely to be awarded in non-coordinated branches. In order to understand these dynamics, we test the coordination \emph{toward} $\Delta$-givers ($G^{\Delta}$). We test the following hypothesis:

\begin{equation}\label{eq:coordination_to_delta_givers}
    \tag{$H_4$}
    C(U, G^{\Delta}) = C(U, G^{\overline{\Delta}})
    \end{equation}

\begin{figure}[h!]
  \includegraphics[width=\columnwidth]{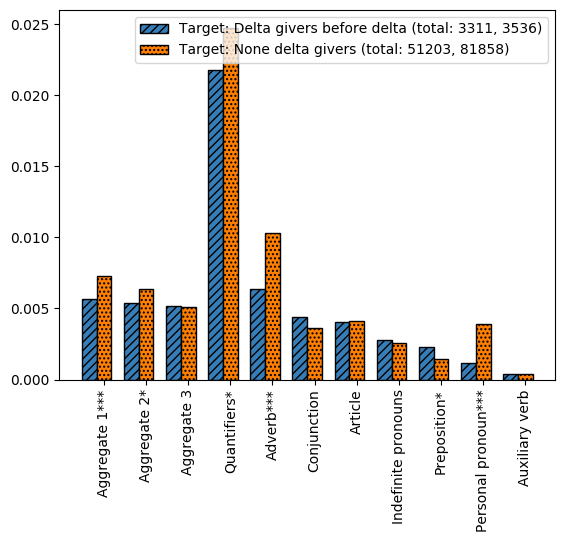}
  \caption{Accommodation towards delta givers vs Non-Delta givers. Test \eqref{eq:coordination_to_delta_givers}}
  \label{fig:target_delta_givers}
\end{figure}

It is important to note that $C$ is calculated only on utterances made \textit{before} the $\Delta$ was awarded. Using the two-sample $t$-test we get mixed results (Figure \ref{fig:target_delta_givers}). Significance is observed for some markers (quantifiers, adverbs, prepositions and personal pronouns) and for Aggregate 1 and 2,  but not for Aggregate 3 and the other markers that proved to be highly distinctive in RQ1-3. 

Following the inconclusive result for $H_4$, we divide $G^{\Delta}$, the group of $\Delta$-givers, to two distinct sets -- OPs ($G^{\Delta_{OP}}$) and $\Delta$-givers that are not OPs ($G^{\Delta_{reg}}$). We can now test whether $\Delta$-giving OPs respond differently than other $\Delta$-givers when \emph{accommodated to}. The idea behind this split is to find out if one group is more susceptible to accommodation than the other. We consider two settings: (i) Accommodation toward non-OP $\Delta$-givers ($G^{\Delta_{reg}}$) vs. accommodation toward non-$\Delta$-givers ($G^{\overline{\Delta}}$), for which we test $H_{4.1}$, and (ii) Accommodation toward $\Delta$-giving OPs ($G^{\Delta_{OP}}$) vs. the accommodation toward OPs that did not award a $\Delta$ ($G^{\overline{OP_\Delta}}$), for which we test $H_{4.2}$. In both settings we calculate the coordination on utterances made \textit{before} the $\Delta$ was awarded, similar to \eqref{eq:coordination_to_delta_givers}.

\begin{equation*}
\label{eq:coordination_to_reg_delta_givers}
    \tag{$H_{4.1}$}
    C(U, G^{\Delta_{reg}}) = C(U, G^{\overline{\Delta}})
\end{equation*}
\begin{equation*}
\label{eq:coordination_to_OP_delta_givers}
    \tag{$H_{4.2}$}
    C(U, G^{\Delta_{OP}}) = C(U, G^{\overline{OP_\Delta}})
\end{equation*}
    
\begin{figure*}[h!]
  \begin{minipage}[t]{0.45\textwidth}
  \vspace{0pt}
  \includegraphics[width=\columnwidth]{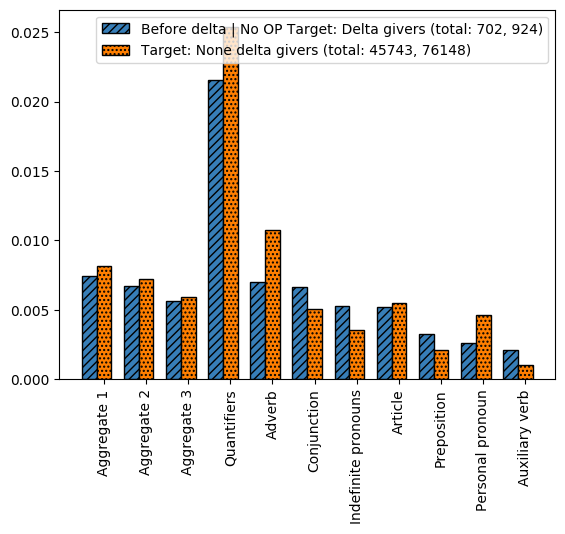}
  \subcaption{Accommodation towards non-OPs that gave delta vs users that did not give Delta. Test \eqref{eq:coordination_to_reg_delta_givers}}
    \label{fig:target_none_OP_delta_givers}
  \end{minipage}\qquad
  \begin{minipage}[t]{0.45\textwidth}
  \vspace{0pt}
  \includegraphics[width=\columnwidth]{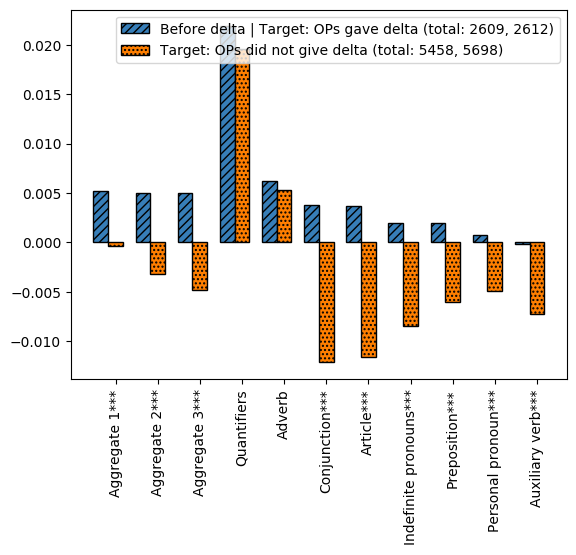}
  \subcaption{Accommodation towards OPs that gave delta vs OPs that did not give Delta. Test \eqref{eq:coordination_to_OP_delta_givers}}
  \label{fig:target_OP_delta_givers}
  \end{minipage}
  \caption{Breaking down RQ4 to two groups}
\end{figure*}

While we cannot reject $H_{4.1}$ (Figure \ref{fig:target_none_OP_delta_givers}), we do reject $H_{4.2}$ (Figure \ref{fig:target_OP_delta_givers}), as significance ($p<0.001$) is observed for all markers and aggregate measures but for adverbs. That is, while regular $\Delta$-givers do not experience any special accommodation, we do observe a dramatic difference in accommodation patterns toward $\Delta$-giving OPs vs. OPs that did not award a $\Delta$ -- OPs are inclined to award a $\Delta$ in discussion threads that are accommodating toward them, and reserve themselves from awarding a $\Delta$ in case the use of style markers diverges over time, rather than coordinated toward them. 

Finally, we complement this analysis by looking at accommodation toward OPs in their own submissions ($G^{OPs}$) and non-OPs ($G^{\overline{OPs}}$). The null hypothesis, asserting that OPs do not experience higher levels of accommodation toward them is formalized as:

\begin{equation*}
\label{eq:coordination_to_OP}
    \tag{$H_{4.3}$}
    C(U, G^{OPs}) = C(U, G^{\overline{OPs}})
\end{equation*}

\begin{figure}[h!]
  \includegraphics[width=\columnwidth]{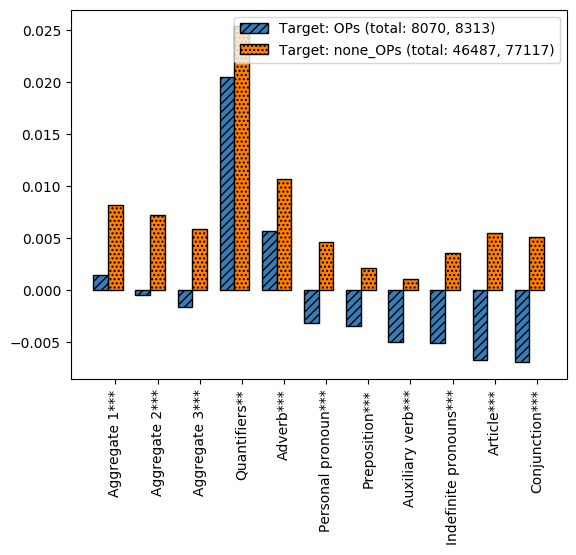}
  \caption{Accommodation towards OPs vs Non-OPs. Test \eqref{eq:coordination_to_OP}}
  \label{fig:to_OPs}
\end{figure}

This hypothesis is rejected with significance of $p<0.001$ for all markers. However, we see that users tend to diverge from the OPs, rather than accommodate toward them (Figure \ref{fig:to_OPs}). This result coupled with the results of  $H_4, H_{4.1}$ and $H_{4.2}$ highlights the importance of accommodation to achieve persuasion, as we briefly discuss on below.

\subsection{Discussion}
\paragraph{Open-mindedness or power and status} Previous work promoted the social framework of power-relations as the force behind linguistic style accommodation. 
Our results suggest that accommodation is related to the willingness of an individual to examine her or his beliefs, consider alternative viewpoints, appreciate convincing argumentation, and ultimately ``convert''. This correlation between open-mindedness and accommodation is observed for OPs and $\Delta$-givers alike -- both present higher accommodation levels \emph{toward} their conversation partners.

\paragraph{Dynamic Open-mindedness} Analysing the accommodation of OPs toward others, we observed that they present significant accommodation in the discussions they initiated, but present accommodation levels similar to other users in the discussions they only take part of as regular users. These results suggest that open-mindedness can be context-dependent, rather than an inherent personal trait. This view could also accommodate (pun intended) previous works promoting the power-structure hypothesis, as the power difference implicitly imposes open-mindedness -- the expected state-of-mind of a subordinate, given our social norms.

\paragraph{Accommodation on the discussion level}
Surprisingly we found that discussion convergence (reflected by a $\Delta$ awarded by the OP) does not positively correlate with accommodation. In fact, $\Delta$-givers that are not OPs are more likely to award the $\Delta$ in discussions that are characterized by style \emph{divergence}, rather than coordination.
We hypothesize that Deltas serve a subtle, secondary, social function, beyond the acknowledgment of a convincing argument. Users may be inclined to award a $\Delta$ in a subconscious effort to ease tensions that are manifested by the divergence of style. This is in line with previous work by \citet{mukherjee2012analysis}, showing that strategic expressions of an agreement by debaters are correlated with high accommodation. Similarly, we observe the ``strategic'' impact of style coordination in promoting agreement -- as OPs are more likely to award a $\Delta$ if they are accommodated to.

\section{Conclusions and Future Work}
\subsection{Conclusions}
We have proposed a novel hypothesis, asserting that linguistic style accommodation correlates with \emph{open-mindedness}. We have provided evidence that supports this hypothesis through a series of experiments. We further demonstrated that open-mindedness is context-dependent, and argued that previous frameworks through which accommodation is studied could be viewed as a special case under the umbrella framework of open-mindedness we have proposed. 

In future work we aim to further explore the different accommodation levels between some of the markers, and the way these are correlated with the conversational discourse structure, as proposed by \cite{zakharov2020discourse}.

Additionally, another area we believe is worth exploring and that is the effect of `performance' as studied by \citet{goffman1967interaction}. OPs, being in a sort of formal role maybe ``acting'' and their accommodation might be altered for it. In this work we used the delta users as a control group that is not affected by this effect. Still, it is interesting to research how accommodation is affected when “acting” compared to genuine behavior. Further exploration of the shifting performative roles of the users, their dynamic roles, their assumed persona should be addressed in future work through a controlled experiment.

\section*{Acknowledgments} We thank the anonymous reviewers for their valuable comments, and especially the suggestion to address accommodation and open-mindedness trough the framework of  `performance'.

\bibliography{eacl_accomodation}
\bibliographystyle{acl_natbib}

\end{document}